\documentclass[letterpaper, 10 pt, conference]{ieeeconf}  % Comment this line out if you need a4paper

\IEEEoverridecommandlockouts                              % This command is only needed if 
\usepackage{graphics} % for pdf, bitmapped graphics files
\usepackage{epsfig} % for postscript graphics files
\usepackage{mathptmx} % assumes new font selection scheme installed
\usepackage{times} % assumes new font selection scheme installed
\usepackage{amsmath} % assumes amsmath package installed
\usepackage{amssymb}  % assumes amsmath package installed
\usepackage{bm}
\usepackage{multirow}
\usepackage{hyperref}
\usepackage{cite}
\usepackage{algorithm}
\usepackage{algorithmic}
\usepackage{url}

\title{\LARGE \bf
Accurate non-stationary short-term traffic flow prediction method
}

\author{Wenzheng Zhao$^{1}$% <-this % stops a space
\thanks{$^{1}$Wenzheng Zhao is from College of Engineering, China Agricultural University,
         Beijing, 100000, China
        {\tt\small zhaowenzheng1999@gmail.com}}%
}

\begin{document}

\maketitle
\thispagestyle{empty}
\pagestyle{empty}
%%%%%%%%%%%%%%%%%%%%%%%%%%%%%%%%%%%%%%%%%%%%%%%%%%%%%%%%%%%%%%%%%%%%%%%%%%%%%%%%
\begin{abstract}

Precise and timely traffic flow prediction plays a critical role in developing intelligent transportation systems and has attracted considerable attention in recent decades. Despite the significant progress in this area brought by deep learning, challenges remain. Traffic flows usually change dramatically in a short period, which prevents the current methods from accurately capturing the future trend and likely causes the over-fitting problem, leading to unsatisfied accuracy. To this end, this paper proposes a Long Short-Term Memory~(LSTM) based method that can forecast the short-term traffic flow precisely and avoid local optimum problems during training. Specifically, instead of using the non-stationary raw traffic data directly, we first decompose them into sub-components, where each one is less noisy than the original input. Afterward, Sample Entropy~(SE) is employed to merge similar components to reduce the computation cost. The merged features are fed into the LSTM, and we then introduce a spatiotemporal module to consider the neighboring relationships in the recombined signals to avoid strong autocorrelation. During training, we utilize the Grey Wolf Algorithm~(GWO) to optimize the parameters of LSTM, which overcome the overfitting issue. We conduct the experiments on a UK public highway traffic flow dataset, and the results show that the proposed method performs favorably against other state-of-the-art methods with better adaption performance on extreme outliers, delay effects, and trend-changing responses.

% In this paper, a short-term traffic flow prediction model is proposed to provide managers with a basis for traffic decision-making to avoid or alleviate traffic congestion. First, the traffic flow data is processed for outliers and missing values. The Complete Ensemble Empirical Mode Decomposition with Adaptive Noise (CEEMDAN) method and Sample Entropy (SE) are used to decompose and reconstruct the non-stationary traffic signal respectively. The purpose is to excavate the information of each component and improve the efficiency of the model. Then, the Grey Wolf Algorithm (GWO) is used to optimize the parameters of the LSTM) in order to improve the prediction accuracy and prevent the model from falling into a local optimum. Finally, The CEEMDAN-SE-GWO-LSTM method proposed in this paper is compared with several conventional methods, including Back PropagationBack Propagation Neural Network (BPNN), Long Short-Term Memory (LSTM), and GWO-LSTM, by using the UK highway traffic flow public data set. The experimental results show that CEEMDAN-SE-GWO-LSTM has a higher prediction accuracy than the compared methods in terms of adapting to extreme values, overcoming delay effects, and stable trend response.

\end{abstract}


%%%%%%%%%%%%%%%%%%%%%%%%%%%%%%%%%%%%%%%%%%%%%%%%%%%%%%%%%%%%%%%%%%%%%%%%%%%%%%%%

\section{Introduction}
With the rapid growth of population and urbanization, traffic congestion has become increasingly severe, leading to social problems such as prolonged travel times and frequent traffic accidents. By leveraging cutting-edge technologies to circumvent traditional infrastructure enhancement constraints, Intelligent Transportation System~(ITS) can effectively ameliorate the congestion and safety issue~\cite{zheng2020traffic, xu2021opencda}. As an essential element in the ITS deployment, traffic flow prediction has attracted significant attention in the research field. By seeing the traffic flow in advance, the government can better allocate traffic resources to reduce congestion, and individuals can make more efficient travel decisions~\cite{zhang2008dynacas}.  

Despite the tremendous potential benefits,  predicting traffic flow accurately and timely is challenging. Traffic flow is usually influenced by various complicated factors such as weather, geography, and the time of day, which are highly nonlinear and volatile. Moreover, traffic forecasts must be capable of foreseeing the traffic in the upcoming future~(typically 15 minutes-30 minutes) to be meaningful, and such a short-term trend is difficult to capture as traffic flow can change dramatically in a short period. Although recent advancement in deep learning has brought remarkable improvement in traffic flow prediction, most existing methods still cannot handle the challenges above well. They either have a slow response to the quick traffic flow change or tend to fall into local optimum during training due to overfitting, leading to unsatisfied accuracy.
% Furthermore, from a macro perspective, traffic flow itself will be affected by external factors (such as urban construction, travel policy, etc.) change over time, which leads to the loss of practical significance for models that rely too much on huge historical data despite their high accuracy.

This paper proposes an ensemble model to solve the short-term traffic flow prediction accuracy. We first use the Complete Ensemble Empirical Mode Decomposition with Adaptive Noise~(CEEMDAN)~\cite{pincus1995approximate} method to decompose the non-stationary signal, where the trend in sub-components is easier to be captured. Meanwhile, to reduce the computational cost, SE is introduced to discard the similar sub-components and only keep the ones with high entropy. The remained components are then fed into LSTM to consider the temporal information. The low-frequency components of the LSTM outputs will be further sent into a Spatiotemporal module to examine the neighboring relationships between them to reduce the autocorrelation. Eventually, the predictions from different components will be aggregated to obtain the final results. During training, we deploy the GWO to optimize the model parameters to avoid the over-fitting issue.

We conduct experiments on a UK public highway dataset to evaluate our approach and compare it with other state-of-the-art models. The results show that our model outperforms other methods both quantitatively and qualitatively.
% Other benchmark models were compared with the model proposed in this paper, and the results have shown that proposed model in this paper has excellent performance in various evaluation indicators. Qualitatively, our proposed model can effectively improve adaptation to extreme values, overcoming delay effects, and stable trend response.

The main contributions of the paper are as follows. 

\begin{itemize}

    \item  We present a robust ensemble model for short-term traffic flow prediction, which can timely and accurately respond to the dramatic change in the traffic flow. Extensive experiments demonstrate that our model can perform better than other baseline methods.
    % Noise reduction of the data is performed by the CEEMDAN-SE method. Due to the volatility and instability of traffic flow, using raw data directly as the input to the model can lead to low prediction accuracy. This paper uses CEEMDAN to decompose the original sequences and use Sample Entropy to measure the complexity of each IMF component. Moreover, this paper combined the subsequences with similar complexities to reduce input dimension and improve prediction efficiency.
    \item We propose a CEEMEAN-SE module to reduce the complexity of the raw traffic data by decomposing it into several sub-components and keeping the computation cost low.
    
    \item We develop a spatiotemporal module to reduce the autocorrelation of low-frequency traffic flow signals.
    
    \item We integrate the GWO optimization in our model, which solves the common over-fitting problem.
    
    % \item   An optimized GWO-LSTM model is built to predict the traffic flow data. Since the traditional LSTM model is prone to fall into local optimum, this paper adopts the GWO algorithm to optimize the parameters of the LSTM, which improves the model's optimization-seeking speed and prediction accuracy. 
    
    % \item   Use spatiotemporal features to reduce the autocorrelation of low-frequency traffic flow signals, establish an integrated optimization model, and improve prediction accuracy.
    
    % \item   Multi-model comparison experiments are constructed. The ensemble model proposed in this paper is compared with BPNN, LSTM, GWO-LSTM, and CEEMDAN-SE-GWO-LSTM. The SSE, MAE, MSE, MAPE, RMSE, and $R^2$ metrics are calculated separately for the above models to quantify prediction accuracy. 
    
\end{itemize}

\section{Related work}

Traffic flow has a strong regularity and periodicity, which is the basis for accurate prediction. However, there is also noticeable uncertainty existing in the short-term traffic flow~\cite{zhou2017delta}, which makes the prediction task challenging. Researchers have devoted themselves to this field in the past years, and the approaches can be divided into parametric and non-parametric methods.

\subsection{Parametric methods}
Autoregressive Integrated Moving Average is a standard parametric method for forecasting time series data, a statistical analysis model that uses time-series data to understand better the data set or predict future trends. Yu et al.~\cite{yu2004switching} propose switching the ARIMA model and applying it to actual data obtained from UTC/SCOOT system. Kumar et al.~\cite{kumar2015short} propose a Seasonal ARIMA model for short-term traffic flow prediction. Chen et al.~\cite{chen2011short} propose an Autoregressive Integrated Moving Average with Generalized Autoregressive Conditional Heteroscedasticity model for traffic flow forecasting. 

Another standard parametric method for time series prediction is the Kalman filter technique. Kumar et al.~\cite{kumar2017traffic} propose a traffic flow prediction model based on the Kalman filter technique. Guo et al.~\cite{guo2014adaptive} propose an Adaptive Kalman filter approach to update the process variances for short-term traffic flow prediction. Although the parametric method has demonstrated effectiveness, it is limited by strong assumptions such as smoothness of the time series, which may lead to poor accuracy when the data varies irregularly in the temporal dimension. Therefore, the parametric approach has limited applicability in the transportation field. 

\subsection{Non-parametric methods}
\subsubsection{Traditional machine learning methods}
Due to the constraints of parametric methods, non-parametric methods have become the first choice for traffic flow prediction nowadays. Yang et al.~\cite{yang2010short} propose a combined wavelet-SVM prediction model for short-term traffic flow prediction. Duan et al.~ \cite{duan2018short} use a particle swarm optimization~(PSO) algorithm to select the appropriate learning parameters to achieve the best PSO-SVM prediction model. Alam et al.~\cite{alam2019prediction} apply five regression models: linear regression, sequential minimum optimization~(SMO) regression, multilayer perceptron, M5P-tree model, and random forest to predict the traffic flow in the city of Porto.

\subsubsection{Deep learning methods}
%  such as cooperative driving automation~\cite{song2021analysis, xu2021opv2v, raju2021evolution, xu2022v2x, chen2022model}
 Recent advancement in deep learning has brought outstanding progress in many fields,  such as autonomous driving~\cite{ xia2022estimation, xu2021holistic, xia2021vehicle, xia2021advancing}, and intelligent transportation system~\cite{raju2021evolution, song2021analysis, song2022federated}. Traffic flow prediction has also been advanced by deep learning. Zhang et al.~\cite{zhang2019short} propose a short-term traffic flow prediction model based on a convolutional neural network~(CNN). Zheng et al.~\cite{zheng2020traffic} propose a traffic flow prediction model based on a Long Short-Term Memory~(LSTM) network. Qu et al.~\cite{zhaowei2020short} propose a new end-to-end hybrid deep learning network model, M-B-LSTM, for short-term traffic flow prediction. Ma et al.~\cite{ma2020daily} use a particular convolutional neural network (CNN) to extract daytime and intra-day traffic flow patterns and feed the extracted features into the LSTM model. Zhao et al.~\cite{zhao2019deep} investigate temporal convolutional networks~(TCN)  for short-term traffic forecasting in the city. Although LSTM is widely used for time series predictions, it tends to be trapped in local optimum when the data is complex and noisy~\cite{wu2021handling}. 

To further reduce noise and improve the prediction accuracy, Chen et al.~\cite{chen2012short} proposed an empirical mode decomposition~(EMD) method. The method decomposes the original short-term traffic flow data into several intrinsic mode functions~(IMFs) and uses them as inputs to the model. However, the IMFs decomposed by the EMD method suffer from mode mixing. Ensemble Empirical Mode Decomposition~(EEMD) improves the mode mixing of EMD by adding Gaussian white noise to the original sequence. Liu et al.~\cite{liu2020short} use EEMD to decompose the traffic flow data into several intrinsic mode functions~(IMFs) and a residue. CEEMDAN improves the processing of EEMD and achieves better decomposition results with higher computational efficiency. Lu et al.~\cite{lu2020hybrid} used the CEEMDAN method to decompose the raw traffic flow data into multiple intrinsic mode function components and a residual component. These methods aim to pre-process the data in a better manner to decrease the data noise. 

\subsection{Traffic flow prediction with perception system}
 As traffic visual recognition is the prerequisite for traffic flow prediction, some researchers also involved perception systems in the field. The key to such system enhancement is whether each vehicle can be accurately recognized. The authors of \cite{chen2020edge} come up with an edge-computing framework that utilizes the real-time multi-object detection and tracking algorithms to provide the inputs to the traffic flow prediction. As cooperative perception system can largely enhance the perceiving range and help see through occlusions~\cite{xu2021opv2v, xu2022v2x, chen2022model},the authors of \cite{li2021cooperative} integrate cooperative perception concept with traffic prediction together to better estimate the traffic states. Similarly, the authors of \cite{kamal2018road} study how to boost the traffic state estimation accuracy by enhancing the perception performance under a partially connected vehicle environment.

% The third section of this paper mainly introduces our method, the fourth section introduces the experimental details and discusses the experimental results, and the fifth section summarizes our research. Finally, this paper discuss the shortcomings and improvements of our scheme.

% It is well known that traffic flow is a signal that varies non-stationarily over time~(frequency varies over time). Previously there are no excellent theories for non-stationary processing signals. The EMD decomposition \cite{huang1998empirical} proposed by Huang NE in 1998, part of the Hilbert-Huang transform (HHT), is a breakthrough in this kind of signal analysis. Huangs' method is based upon spectral decomposition. Spectral decomposition decomposes the signal into components of different frequencies.

\section{Methodology}

We propose a robust ensemble model that can effectively learn from non-stationary signals and utilizes spatiotemporal features to optimize the traffic prediction process, avoiding common problems including over-fitting, local optimum, and autocorrelation. 

\begin{figure*}[htbp]
    \centerline{\includegraphics[width=1.0\textwidth]{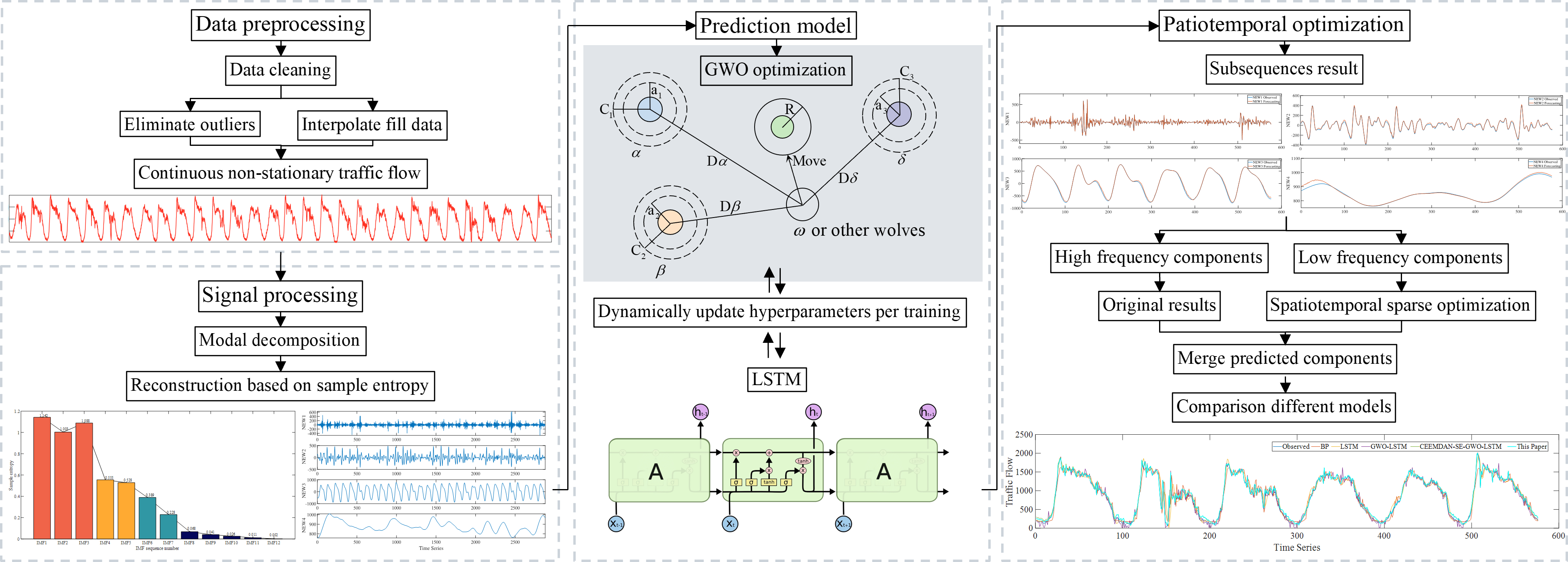}}    \caption{The pipeline of our traffic prediction approach. }
   \label{fig:1}
\end{figure*}

We first adopt the theory of CEEMDAN~\cite{torres2011complete} to decompose the original non-stationary traffic flow signals into several IMF sub-signals with different frequencies. We then utilize SE~(sample entropy) to measure the nonlinear complexity of the IMF subsequence to merge similar sub-signals for computation reduction. For the prediction model, we deploy the LSTM model with GWO optimization to improve the prediction accuracy and decrease the training duration. Finally, we will extract the spatiotemporal features from the output of LSTM to avoid the autocorrelation problem. The pipeline of our approach is demonstrated in Fig.~\ref{fig:1}. 

\subsection{CEEMDAN-SE}

\subsubsection{CEEMDAN}

EMD~(Empirical Mode Decomposition) is a classic adaptive method for solving non-stationary signal problems~\cite{huang1998empirical}. This method utilizes the signal extreme point information to decompose the signal into several IMFs~(Intrinsic Mode Functions). However, the modal aliasing problem of EMD will cause severe sawtooth lines in the time-frequency distribution and makes certain eigenmode functions lose their physical meaning, which leads to the degradation of the performance of EMD. 

%Based on the CEEMD-SE method, Keke Wang et al. \cite{wang2019wind} propose a wind power short-term prediction model. In their experiment, the RMSE and MAE are 2.16 and 0.39, respectively, better than EMD-SE-HS-KELM, HS-KELM, KELM, and ELM models. In 2020, Tian Z \cite{tian2020approach} propose a short-term traffic flow prediction model. The model is based upon empirical mode decomposition and combination model; their proposed model demonstrate superior performance to the state-of-the-art prediction methods or models in short-term traffic flow prediction. However, the two models have limitations. Neither model takes noise into account; neither can automatically adjust to follow changes in noise. 

\begin{figure}[htb]
    \centerline{\includegraphics[width=0.5\textwidth]{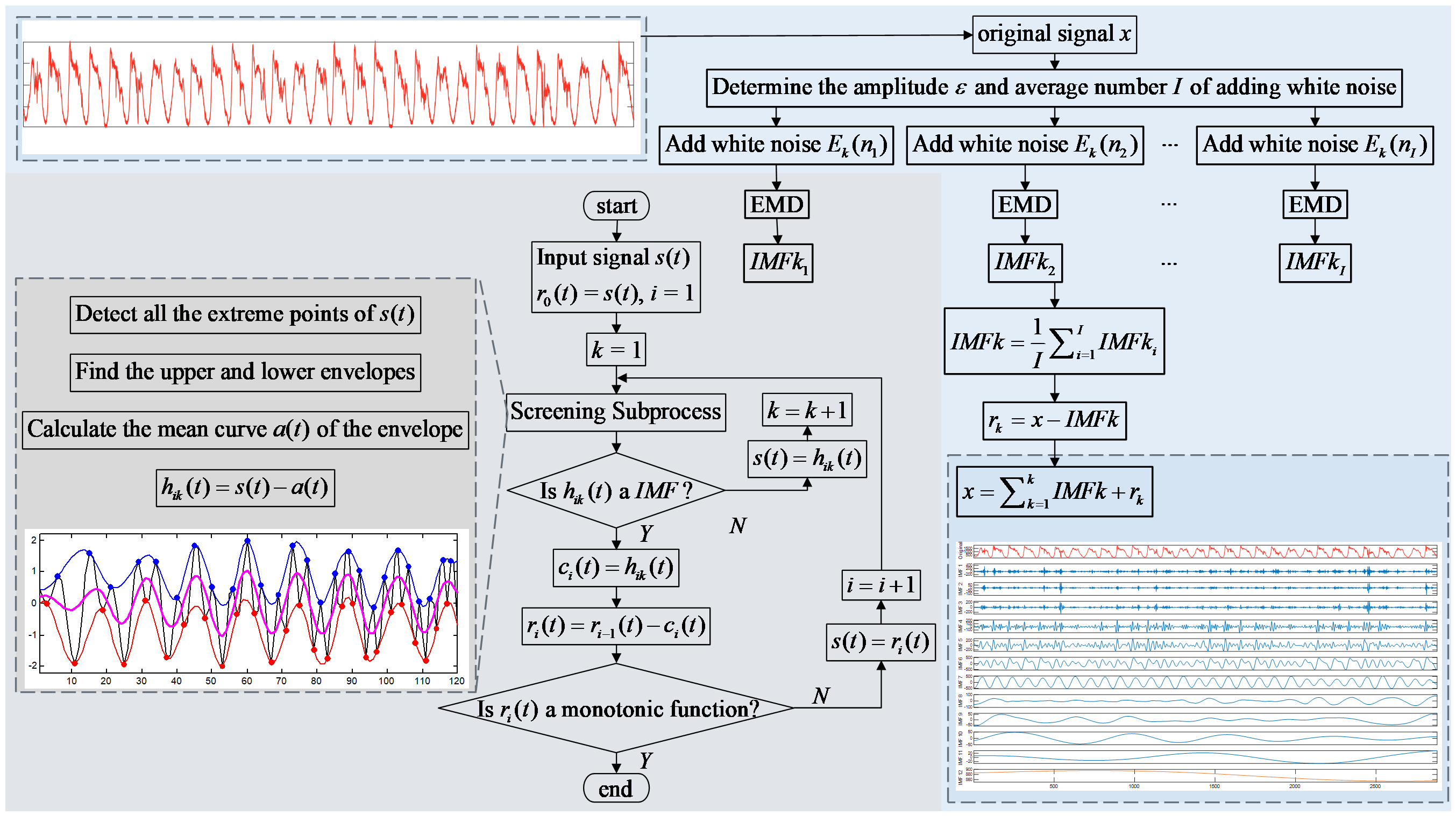}}
    \caption{The process of the CEEMDAN algorithm. }
   \label{fig:2}
\end{figure}

%Therefore, We propose the CEEMDAN (Complete Ensemble Empirical Mode Decomposition with Adaptive Noise) \cite{torres2011complete} method to decompose the traffic flow time series to reduce its non-stationarity. 
Based on EMD, we develop the CEEMDAN~(Complete Ensemble Empirical Mode Decomposition with Adaptive Noise) algorithm, which applies the EEMD~(Ensemble Empirical Mode Decomposition) method to add Gaussian noise to the original signal. The mode mixing problem is solved by stacking and averaging multiple operations to cancel the influence of noise to gain better mode decomposition results. The process of the CEEMDAN algorithm is shown in Fig.~\ref{fig:2}. 

\subsubsection{Sample Entropy Theory}

$N$ sub-sequences will be generated after the CEEMDAN decomposition. Directly using them as the input data of the prediction model GWO-LSTM will cause a large computational cost. Therefore, we employ SE~(sample entropy)~\cite{richman2000physiological}, an approximate entropy~(ApEn)~\cite{pincus1995approximate} method that evaluates time series complexities by measuring the probability of the new generated patterns, to classify and reconstruct the traffic flow temporal data for reducing the complexity of the sub-sequences.  SE is defined as the negative natural logarithm of the conditional probability, where self-matches are not included:

\begin{equation}
    \text{SampEn}(m, r)=\lim _{N \rightarrow \infty}\left\{-\ln \left[\frac{A^{m}(r)}{B^{m}(r)}\right]\right\}
    \label{eq:1}
\end{equation}

${B^{m}(r)}$ in Eq.~\eqref{eq:1} is the probability of the two sequences matching $m$ points under the similarity tolerance $r$, and ${A^{m}(r)}$ is the probability of the two sequences matching $m+1$ points. The calculation formulas are Eq.~\eqref{eq:4} and Eq.~\eqref{eq:5}, respectively. 
  
\begin{equation}
    A_{i}^{m}(r)=\frac{1}{N-m-1} A_{i}
    \label{eq:2}
\end{equation}

\begin{equation}
    B_{i}^{m}(r)=\frac{1}{N-m-1} B_{i}
    \label{eq:3}
\end{equation}

\begin{equation}
    A^{m}(r)=\frac{1}{N-m} \sum_{i=1}^{N-m} A_{i}^{m}(r)
    \label{eq:4}
\end{equation}

\begin{equation}
    B^{m}(r)=\frac{1}{N-m} \sum_{i=1}^{N-m} B_{i}^{m}(r)
    \label{eq:5}
\end{equation}
  
${A_{i}}$ and ${B_{i}}$  are the number of the maximum distance, not greater than $r$, between the vector sequences  ${X_{m}(i)}$  and  ${X_{m}(j)}$ of the dimension $m$ composed of time series data when the dimension is $m+1$ and $m$ respectively. Specifically, $X_{m}(i)=\{x(i), x(i+1), \ldots, x(i+m-1)\}$ , $1 \leq i \leq N-m+1$ , represents $m$ consecutive values of $x$ starting from the $i$th point. 

The amount of data is usually limited in specific applications. Thus, Eq.~\eqref{eq:1} evolved into Eq.~\eqref{eq:6}. 

\begin{equation}
    \operatorname{SampEn}(m, r, N)=-\ln \left[\frac{A^{m}(r)}{B^{m}(r)}\right]
    \label{eq:6}
\end{equation}

\subsection{GWO-LSTM}

LSTM~\cite{hochreiter1997long} is first presented to solve complex artificial long-time-lag tasks. As the GWO algorithm~\cite{prasanth2021forecasting} mimics the leadership hierarchy and hunting mechanism of grey wolves in nature, it could be used to optimize the LSTM model. Our model has a significant optimization effect compared to the state-of-the-art LSTM neural networks and BP neural networks. Thus, we use the GWO-LSTM method to complete the post-processing of CEEMDAN-SE. The general process of the GWO-LSTM framework part is shown in Fig.~\ref{fig:1}. 

%\begin{figure*}[htb]
%    \centerline{\includegraphics[width=1.0\textwidth]{figs/GWO-LSTM-chart.png}}
%    \caption{The presented GWO-LSTM framework. }
%   \label{fig:3}
%\end{figure*}

\subsubsection{LSTM}

Recurrent neural network~(RNN)~\cite{elman1990finding} cannot solve the long-term dependence problem in which the output is related to a long sequence of preceding segments. Thus, LSTM is designed to solve this problem. Compared to RNN, LSTM has three more gates - forgetting gate, input gate, and output gate - enabling it to achieve better results in traffic flow prediction. 

Since the output is a linear combination of the inputs, the nonlinearity of LSTM needs to be enhanced. The enhancement will be done through the use of the activation function as it exacerbates the nonlinearity of the network model. Common activation functions for LSTM are tanh(-1, 1), sigmod (0, 1) ,and relu[0, 1). Following experimental verification, tanh(-1, 1) presents better results to our problem and is selected as our activation function. 

\subsubsection{GWO}

We first divide the traffic flow prediction into four layers and enter them into the GWO model to complete the initialization, with the first three layers being of greater significance. We define $\alpha$ as the optimum solution. During the hunt, the behavior of grey wolves rounding up their prey is defined as Eq.~\eqref{eq:7} and Eq.~\eqref{eq:8}, where $t$ is the current iterative generation, $\bm{A}$ and $\bm{C}$ are the coefficient vectors, $\bm{X}_p$ and $\bm{X}$ are the prey position vector and the grey wolf position vector, respectively. 

\begin{equation}
    \bm{D}=\left|{\bm{C}\bm{X}}_{p}(t)-\bm{X}(t)\right|
    \label{eq:7}
\end{equation}

\begin{equation}
    \bm{X}(t+1)=\bm{X}_{p}(t)-{\bm{A}\bm{D}}
    \label{eq:8}
\end{equation}

The calculation equations of $A$ and $C$ are shown in Eq.~\eqref{eq:9} and Eq.~\eqref{eq:10}, where $\alpha$ is the convergence factor. As the number of iterations decreases linearly from 2 to 0, the norms of $r_1$ and $r_2$ are random numbers between [0, 1]. 

\begin{equation}
    \bm{A}=2 \alpha r_{1}-\alpha
    \label{eq:9}
\end{equation}

\begin{equation}
    \bm{C}=2 r_{2}
    \label{eq:10}
\end{equation}

In the GWO model, the upper layer leads the lower layer to the set of update equations shown in Eq.~\eqref{eq:11}, and after completing the update, the GWO model outputs $\bm{X}(t+1)$ to the LSTM model according to Eq.~\eqref{eq:12}. Subsequently, the model calculates the loss function and adjusts the learning rate of the GWO model according to the vector $\bm{X}$. 

\begin{equation}
    \left\{
        \begin{array}{l}
        {D}_{\alpha}=\left|{C}_{1}\bm{X}_{\alpha}-\bm{X}\right|, \\
        {D}_{\beta}=\left|{C}_{2}\bm{X}_{\beta}-\bm{X}\right|, \\
        {D}_{\delta}=\left|{C}_{3}\bm{X}_{\delta}-\bm{X}\right|
        \end{array}
    \right. 
    \label{eq:11}
\end{equation}

\begin{equation}
    {X}(t+1)=\frac{\left({X}_{\alpha}-{A}_{1} {D}_{\alpha}\right)+\left({X}_{\beta}-{A}_{2} {D}_{\beta}\right)+\left({X}_{\delta}-{A}_{3} {D}_{\delta}\right)}{3}
    \label{eq:12}
\end{equation}

\subsubsection{Combining GWO with LSTM}

We reference the data on the LSTM model to derive a prediction of the baseline model. Subsequently, We incorporate the LSTM prediction results into the GWO model to obtain the new four strata. Once the four strata are obtained, GWO will calculate the coefficient matrices $\bm{A}$ and $\bm{C}$ according to Eq.~\eqref{eq:9} and Eq.~\eqref{eq:10}. Then it will calculate the ratios of each stratum in the four strata using $\bm{A}$ and $\bm{C}$, inputting them into the LSTM model for automated parameter tuning, and continue to train the LSTM model. The above process will then repeat until a user-specified number of iterations is reached. 

Machine learning training aims to update the parameters and optimize the objective function. In this paper, the GWO is set as an optimizer to perform a local estimation~($|\bm{A}|$ shown as Eq.~\eqref{eq:9}) based on the results of each LSTM iteration to minimize the loss function. We use 1024 as the initial batch size and 0.01 as the initial learning rate. GWO decides to update or not update~(eliminate or not eliminate) the population of grey wolves based on the results of each LSTM iteration and, thus, dynamically adjusts the learning rate of the LSTM each time. In addition, our GWO network is optimized for four layers. 

\subsection{Spatiotemporal optimization for low frequencies IMFs}

Due to the obvious daily cycle characteristics of traffic flow, training with a small amount of data can easily cause autocorrelation problems, resulting in low instantaneous accuracy. Directly training undecomposed instability short-term traffic flow data will lead to the problems mentioned above, and at the same time, the decomposed traffic flow has the problem of data scale difference. Accurate predictions on low frequency components often determine the overall performance of the model. Therefore, we propose to consider spatial characteristic factors in the low-frequency synthetic component~(larger value, representing macroscopic changes). Taking the traffic flow of neighboring stations at the previous moment as input can increase the feature dimension and avoid the autocorrelation problem. The algorithm details are shown in Algorithm~\ref{alg1}.

\begin{algorithm}
	%\textsl{}\setstretch{1.8}
	\renewcommand{\algorithmicrequire}{\textbf{Input:}}
	\renewcommand{\algorithmicensure}{\textbf{Output:}}
	\caption{Spatiotemporal spare optimization}
	\label{alg1}
	\begin{algorithmic}[1]
	    \STATE Given $\left\{ {S_i} \right\}$ $\leftarrow$ Target site and its adjacent traffic flow
		\STATE Initialization: Extract low frequency components $\left\{ {Y_i} \right\}$ of each $\left\{ {S_i} \right\}$, $n \leftarrow 0$
		\STATE Pre-training Spatiotemporal-LSTM with $\left\{ {Y_i} \right\}$
		\REPEAT
		\STATE $n \leftarrow n + 1$
		\STATE Update $\hat{y}^i_{1}$ based on GWO-LSTM
		\STATE Update $\hat{y}^i_{2}$ based on Spatiotemporal-LSTM
		\STATE $error_{t-1}$ = $min{||error^i_{t-1}||}_{2}$
		\STATE Update $\hat{y}^i_{t}$ based on $error_{t-1}$
		\UNTIL End of sub-sequence
		%\UNTIL $\sum\limits_{k=1}^P  {{{\left\| {s_{k,t}^{n + 1}\left( \omega  %\right) - s_{k,t}^n\left( \omega  \right)} \right\|_2^2} \mathord{\left/
		%			{\vphantom {{\left\| {s_{k,t}^{n + 1}\left( \omega  \right) - s_{k,t}^n\left( \omega  \right)} \right\|_2^2} {\left\| {s_{k,t}^n\left( \omega  \right)} \right\|_2^2}}} \right.
		%			\kern-\nulldelimiterspace} {\left\| {s_{k,t}^n\left( \omega  \right)} \right\|_2^2}}}  < \varepsilon $  
		\ENSURE  Low frequency signal predicted value $\hat{Y}_{i}$
	\end{algorithmic}  
\end{algorithm}

Finally, we construct an integrated model for short-term traffic flow prediction based on unsteady signal decomposition and optimization of spatiotemporal features, which can better adapt to instantaneous traffic changes, overcome delayed response, and avoid overfitting caused by small data samples, etc.

\section{Experiment}

\subsection{Dataset}

%Among the most significant ways of travel, flight remains one of the most prevalent means by which the public opts. The geographical location of an airport generally is distant from an individual’s area of residence. Hence, providing accurate traffic information is essential for planning. We utilize the measured traffic flow data of high-speed stations near the M25 Heathrow Airport\cite{timmurphy.org} to ensure the reliability and authenticity of predicated results. The data set is the traffic flow data for 30 consecutive days from September 1, 2019, to September 30, 2019, with a collection frequency of 15 minutes. The data volume became 2880 at each site ensuing the interpolation method. The interpolation method is applied to fill in the missing values of the time series and remove the outliers. We uniformly divide the dataset into a training set and test set according to the ratio of 8:2.

We conduct experiments on the British Highways dataset~\cite{timmurphy.org}, which is published and maintained by the British Highways Agency. This dataset contains the majority of highways in British, and the collection frequency of traffic flow at each highway station is 15 minutes. As the traffic flows near the airports are usually challenging for traffic estimation methods , we majorly compare different approaches on the M25 motorway near Heathrow Airport~(a subset of the British Highways dataset). The chosen subset covers detailed information for the traffic states, including Day Type~(working day or special day), Vehicle Flows~(The number of difference length vehicles detected on any lane within the 15-minute time slice), Speed~(The average speed in km/h over the 15-minute period), and Quality Index~(the number of valid one minute reported and used to generate the Total Traffic Flow and speed). The data volume became 2880 at each site ensuing the interpolation method, which is applied to fill in the missing values of the time series and remove the outliers. The ratio between the samples of training and testing set is 8:2.

%\subsection{CEEMDAN-SE}

%We first employ the CEEMDAN method to decompose the traffic flow. The decomposition result is shown in Fig.~\ref{fig:4}. After the CEEMDAN decomposition, We obtain the first 11 IMF components with different complexities and one IMF residual component with a relatively gentle change.

\subsection{Evaluation Metrics}

% 定量+定性 评估为未来15分钟的交通流量精度。
% 评价函数 定性六个指标 + 分析3方面定性 ：extreme outliers, delay effects, and trend-changing responses
% 实施细节 CEEMDAN-SE -> GWO-LSTM -> Patiotemporal optimization at low frequences sub-signal
% compare models : BP LSTM GWO-LSTM CEEMDAN0-SE-GWO-LSTM 

\begin{table*}[htb]
    \caption{The results of different models. }
    \centering
    \begin{tabular}{ccccccc}
        \hline
        \textbf{Model} & \textbf{SSE} & \textbf{MAE} & \textbf{MSE} & \textbf{RMSE} & \textbf{MAPE} & \textbf{\bm{$R^2$}}  \\
        \hline
        BP & 11226551 & 93.6000 & 19524.4368 & 139.72987 & 0.16351147 & 0.926  \\
        LSTM & 9756640 & 79.9905 & 16938.6070 & 130.1484 & 0.1513894 & 0.936  \\
        GWO-LSTM & 3560689 & 43.4375 & 6181.7526 & 78.6241 & 0.112324 & 0.977  \\
        CEEMDAN-SE-GWO-LSTM & 1244319 & 37.3297 & 2160.2774 & 46.4788 & 0.106236 & 0.992  \\
        \textbf{Our method} & \textbf{1029780} & \textbf{32.4757} & \textbf{1787.8139} & \textbf{42.2825} & \textbf{0.094795} & \textbf{0.993}  \\
        
        \hline
    \end{tabular}
    \label{tab:3}
\end{table*}
The total duration of the selected dataset is one month, and the data in the last week in this month belongs to the test set. We predict the traffic flow every 15 minutes to assess whether the model can capture the traffic changes timely. To quantitatively evaluate the performance of the models, six commonly evaluation metrics are utilized in this paper: Sum of squares error~(SSE), Mean Square Error~(MSE), Root Mean Square Error~(RMSE), Mean Absolute Error~(MAE), Mean Absolute Percentage Error~(MAPE), and R-Squared~($R^2$). The smaller number on these metrics represent better model performance. We compare our methods with Back Propagation neural network~(BP), the vanilla LSTM, GWO-LSTM without signal input signal preprocessing, and CEEMDEAN-SE-GWO-LSTM that does not consider spatiotemporal correction.
% In the qualitative analysis we focus on discussing three aspects that have a significant impact on the 
% prediction of traffic conditions: extreme outliers, delay effects, and trend-changing responses, since poor performance in these three aspects may cause the collapse of the traffic system.

\begin{figure}[htb]
    \centerline{\includegraphics[width=0.5\textwidth]{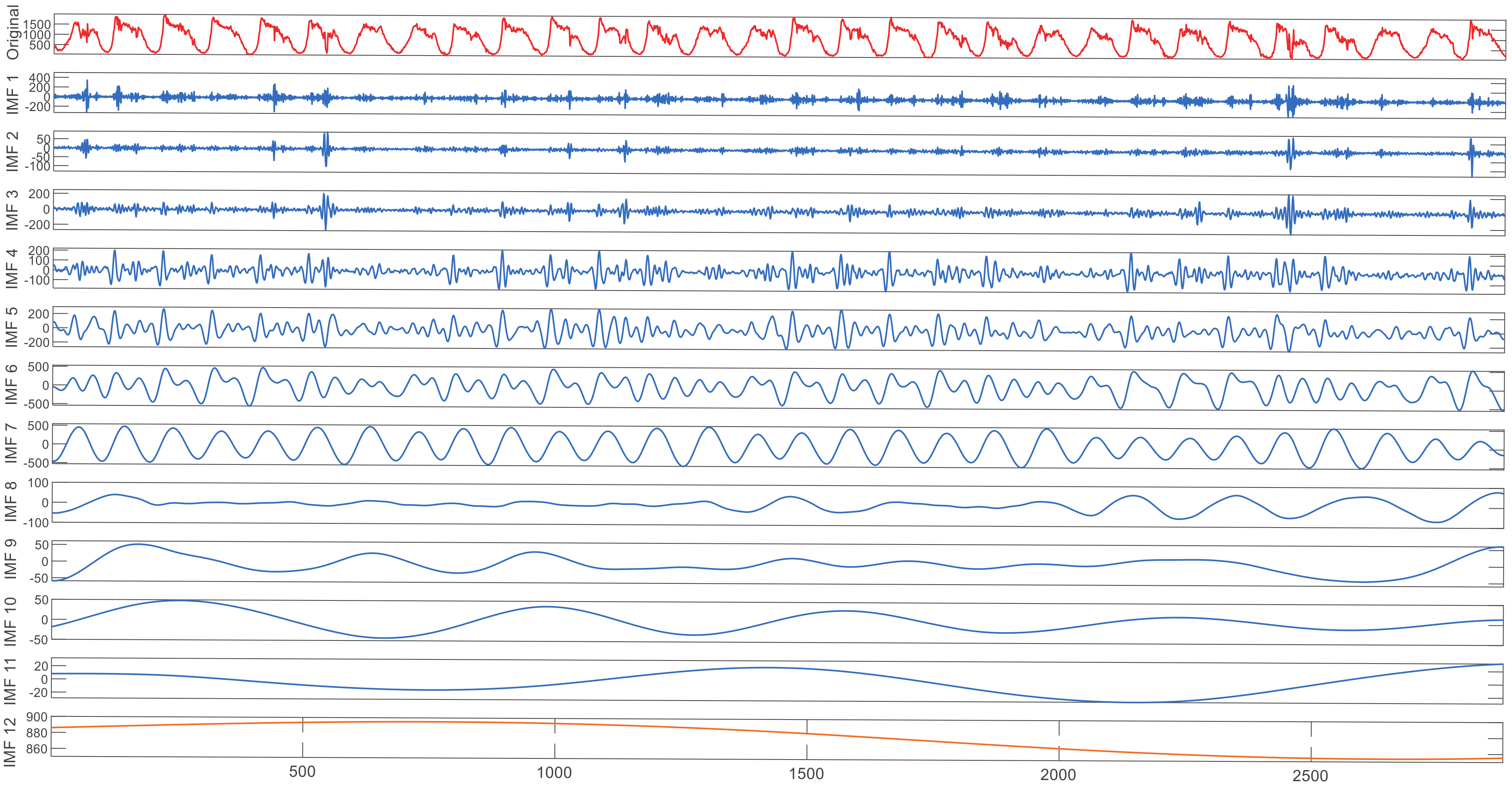}}
    \caption{The decomposition results of CEEMDAN. }
    \label{fig:4}
\end{figure}

\subsection{Implementation Details}
\begin{figure}[htb]
    \centerline{\includegraphics[width=0.5\textwidth]{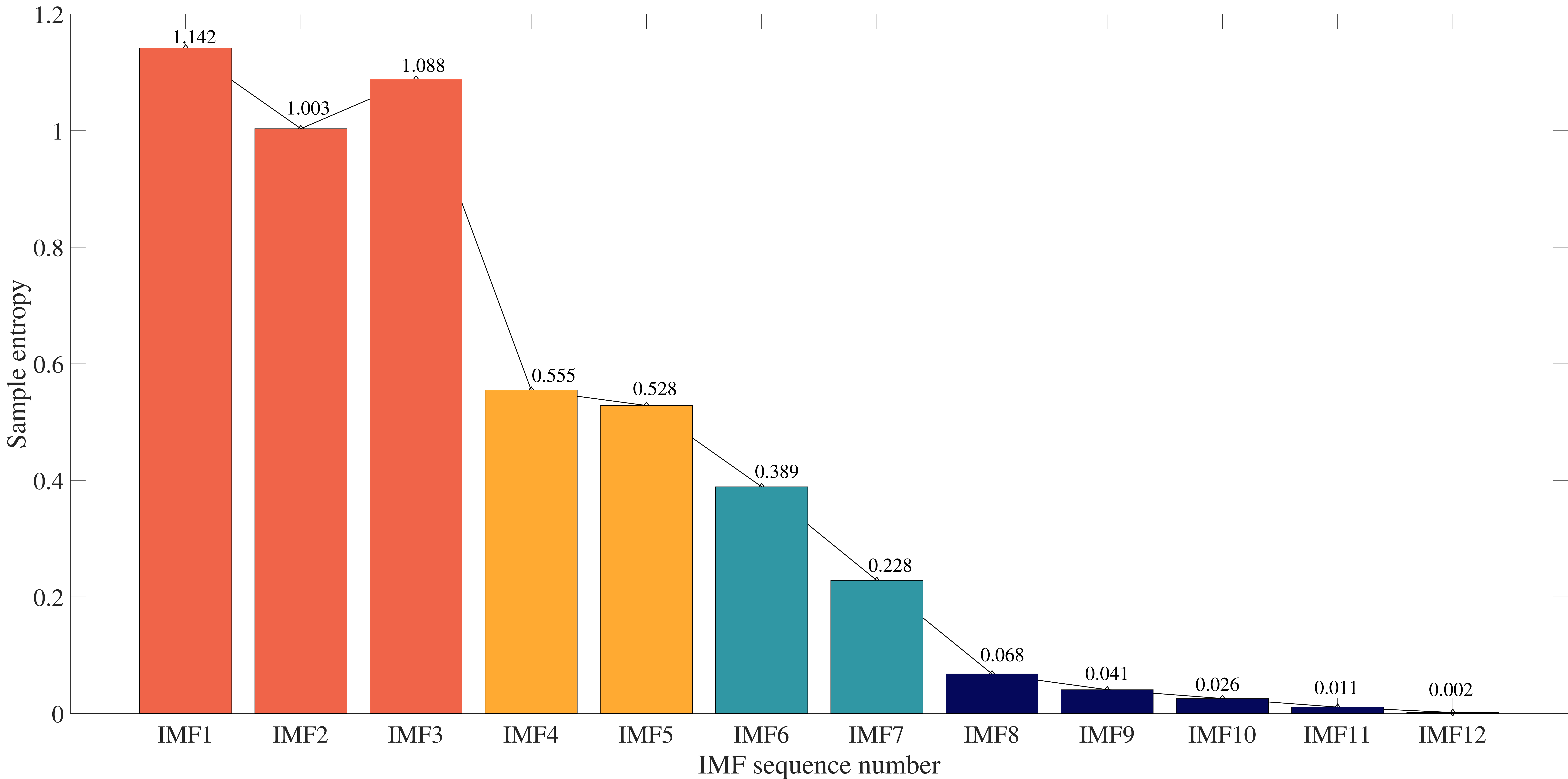}}
    \caption{The sample entropy of each sub-sequence. }
    \label{fig:5}
\end{figure}

\begin{figure}[htb]
    \centerline{\includegraphics[width=0.5\textwidth]{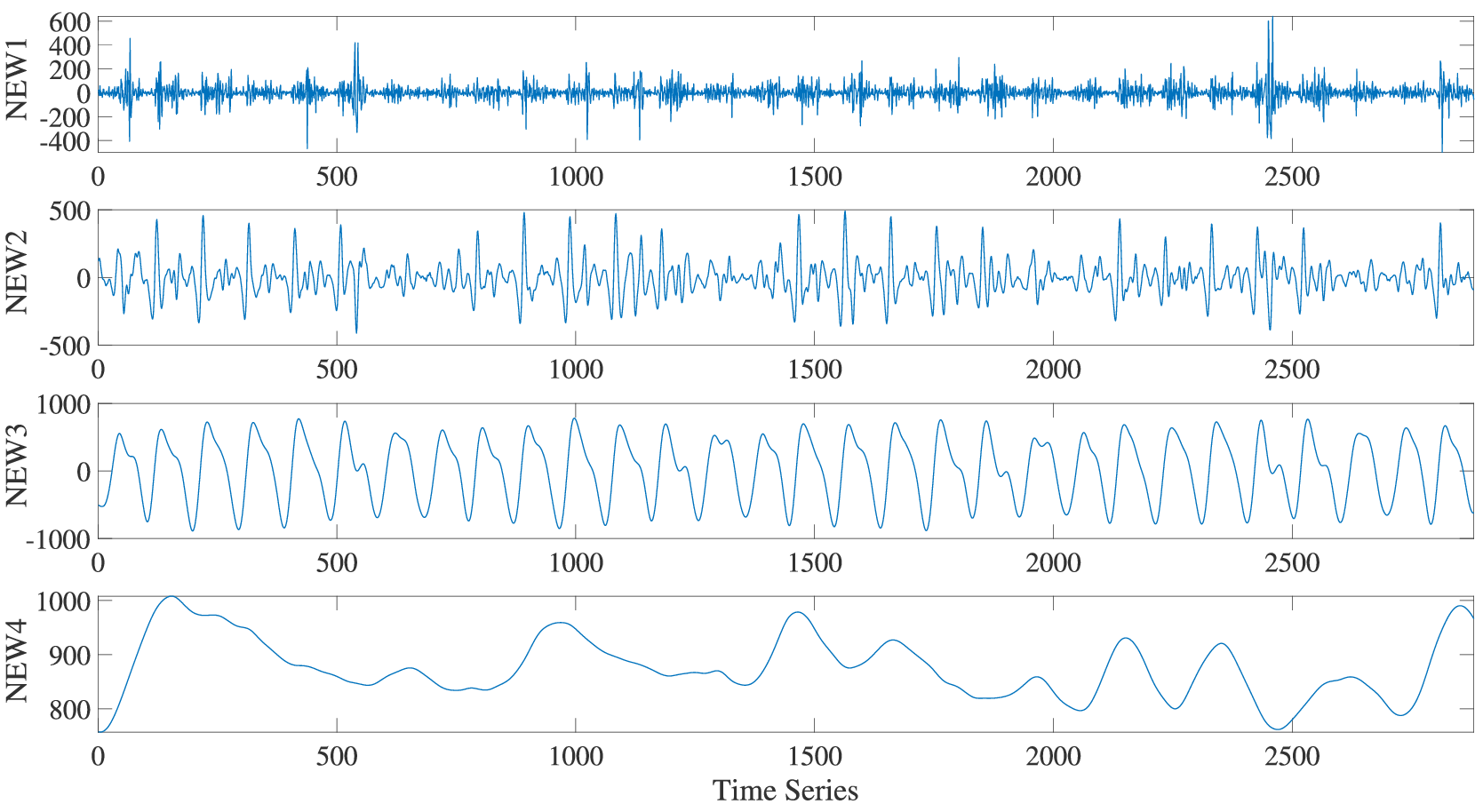}}
    \caption{The reconstitution components of traffic flow through CEEMDAN-SE. }
    \label{fig:6}
\end{figure}

Before feeding the data into the model, we first decompose the original non-stationarity one-dimensional traffic flow signal into 12 IMF sub-signals as shown in Fig.~\ref{fig:4}, In the process of empirical mode decomposition, we set the standard deviation of the noise as 2000, the number of realization to 500, and the maximum number of sifting iterations to 2000. In order to reduce the computational cost, some of the IMF components are merged. For example, IMF1, IMF2, and IMF3 have similar complexities (as shown in Fig.~\ref{fig:5}), and thus they are combined into a single component. The details of the recombination results are shown in Table~\ref{tab:1}, and the new subsequence after reconstruction is shown in Fig.~\ref{fig:6}. We take the above processed data as the input to the GWO-LSTM model, with a batch size of 1024, an initial learning rate of 0.01, and a time window size of 3.  All models are trained with 200 iterations on a Quadro P1000.

%We introduce the theory of sample entropy. We introduce said theory in order to group and reconstruct the IMF sub-signals that are obtained after decomposition. Fig.~\ref{fig:5} is the sample entropy of the IMF component. The IMF sub-functions of similar sample entropy are superimposed to reduce the input sample size of GWO-LSTM.

%Analysis of Fig.~\ref{fig:5} demonstrated that, as the frequency of each IMF component decreases, the corresponding SE value also decreases. In this paper, the entropy value of each component is used as the judging standard; each IMF component is reorganized. Specifically, IMF1, IMF2, and IMF3 have similar complexities and thus can be combined into recombined components. Similarly, the remaining IMF components are merged and recombined, the recombination result is shown in Table~\ref{tab:1}, and the new subsequence after reconstruction is shown in Fig.~\ref{fig:6}. 

\begin{table}[htb]
    \caption{The reconstructed subordinate list. }
    \centering
    \begin{tabular}{cccc}
        \hline
        \textbf{IMFn} & \textbf{Hse(n)} & \textbf{Merged sequence} & \textbf{New component}  \\
        \hline
        IMF1 & 1.142 & IMF1, IMF2, IMF3 & NEW1  \\
        IMF2 & 1.003 &  &   \\
        IMF3 & 1.088 &  &   \\
        IMF4 & 0.555 & IMF4, IMF5 & NEW2  \\
        IMF5 & 0.528 &  &   \\
        IMF6 & 0.389 & IMF6, IMF7 & NEW3  \\
        IMF7 & 0.228 &  &   \\
        IMF8 & 0.068 & IMF8 - IMF12 & NEW4  \\
        IMF9 & 0.041 &  &   \\
        IMF10 & 0.026 &  &   \\
        IMF11 & 0.011 &  &   \\
        IMF12 & 0.002 &  &   \\
        \hline
    \end{tabular}
    \label{tab:1}
\end{table}

%\subsection{GWO-LSTM}

%Experiments are conducted based on the above data to verify the effectiveness of the GWO-LSTM model on traffic flow prediction. In this paper, the original samples are divided into a training set and a testing set according to the ratio of 8:2. The initialized batch size is 1024, and the learning rate is 0.01. The number of hidden layers of the LSTM network chosen is two. The hyperparameters of the LSTM model are continuously and dynamically adjusted in the iterative process using the GWO algorithm with four implied layers until the loss function is obtained below a predefined specific value.  

%\subsection{Patiotemporal optimization for low frequencies IMFs}

%For low-frequency components, in order to avoid low accuracy caused by autocorrelation, modal decomposition is performed on the contemporaneous traffic flow signals of adjacent stations, and components with similar sample entropy are superimposed.The low-frequency components of the site to be predicted and the neighboring sites are taken as input samples, and the length of the time window is reduced to form a spatiotemporal LSTM model, and its results are compared with the GWO-LSTM results. The result with smaller error is selected as the prediction result to improve the prediction performance.
\subsection{Quantitative Results}
Table~\ref{tab:3} shows the quantitative results of different models, and it can be seen that our proposed approach method outperform all others. Our proposed model is 90.8\%, 65.3\%, 90.8\%, 69.7\%, and 42\% lower in  SSE, MAE, MSE, RMSE, and MAPE evaluation compared to the BP model, and 7.2\% higher in the $R^2$. Similarly, compared to LSTM, these numbers are 89.4\%, 59.4\%, 89.4\%, 67.5\%, 37.4\% and 6\% respectively; and those values for GWO-LSTM are 71.1\%, 25.2\%, 71.1\%, 46.2\%, 15.6\% and 1.6\% while for CEEMDAN-SE-GWO-LSTM they are 17.2\%, 13\%, 17.2\%, 9\%, 10.7\% and 0.1\% respectively.

\subsection{Qualitative Analysis}

%Six commonly used error evaluation indexes are used to evaluate the model in this paper. The related equations of the six error evaluation indexes mentioned are shown from Eq.~\eqref{eq:13} to Eq.~\eqref{eq:18}, where $n$ is the number of test set data, and $y_i$  and  $\hat{y}_{i}$ are the actual traffic flow value and predicted traffic flow value at moment $i$ respectively. The smaller the values of MAPE, RMSE, MSE, SSE, and MAE, the smaller the error and the better the prediction effect. The larger $R^2$, the better the prediction effect of the model.

%\begin{equation}
%    {MAPE}=\frac{1}{n} \sum_{i=1}^{n}\left|\frac{\hat{y}_{i}-y_{i}}{y_{i}}\right| \times 100 \%
%    \label{eq:13}
%\end{equation}

%\begin{equation}
%    {RMSE}=\sqrt{\frac{1}{n} \sum_{i=1}^{n}\left(\hat{y}_{i}-y_{i}\right)^{2}}
%    \label{eq:14}
%\end{equation}

%\begin{equation}
%    {MSE}=\frac{1}{n} \sum_{i=1}^{n}\left(\hat{y}_{i}-y_{i}\right)^{2}
%    \label{eq:15}
%\end{equation}

%\begin{equation}
%    {SSE}=\sum_{i=1}^{n}\left(\hat{y}_{i}-y_{i}\right)^{2}
%    \label{eq:16}
%\end{equation}

%\begin{equation}
%    {MAE}=\frac{1}{n} \sum_{i=1}^{n}\left|\hat{y}_{i}-y_{i}\right|
%    \label{eq:17}
%\end{equation}

%\begin{equation}
%    R^{2}=1-\frac{\sum_{i=1}^{n}\left(\hat{y}_{i}-y_{i}\right)^{2}}{\sum_{i=1}^{n}\left(\bar{y}%_{i}-y_{i}\right)^{2}}
%    \label{eq:18}
%\end{equation}

\begin{figure*}[htb]
    \centerline{\includegraphics[width=1.0\textwidth]{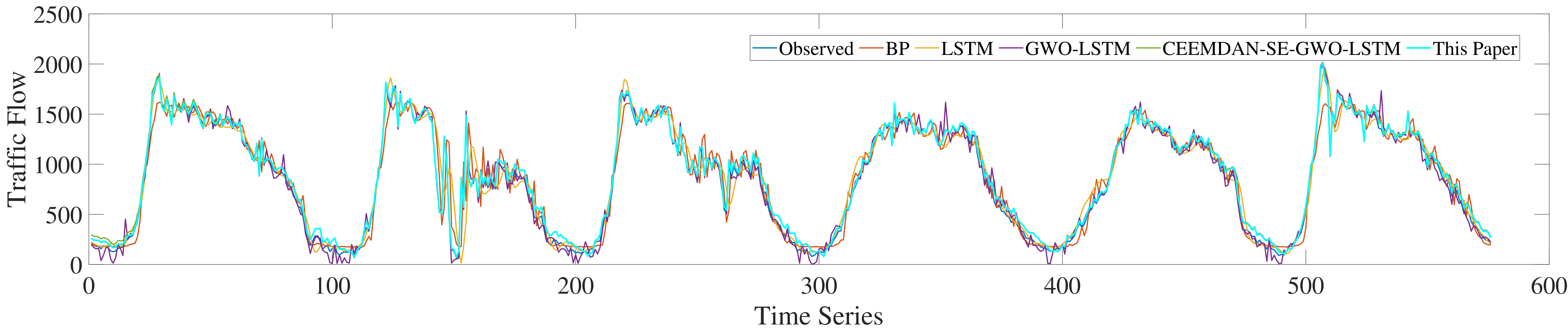}}
    \caption{Comparison of different models. }
    \label{fig:Comparison of different models}
\end{figure*}

%\begin{figure}[htb]
%    \caption{The results of the BP, LSTM, and GWO-LSTM models . }
%    \label{fig:7}
%\end{figure}

\begin{figure}[htb]
    \centerline{\includegraphics[width=0.5\textwidth]{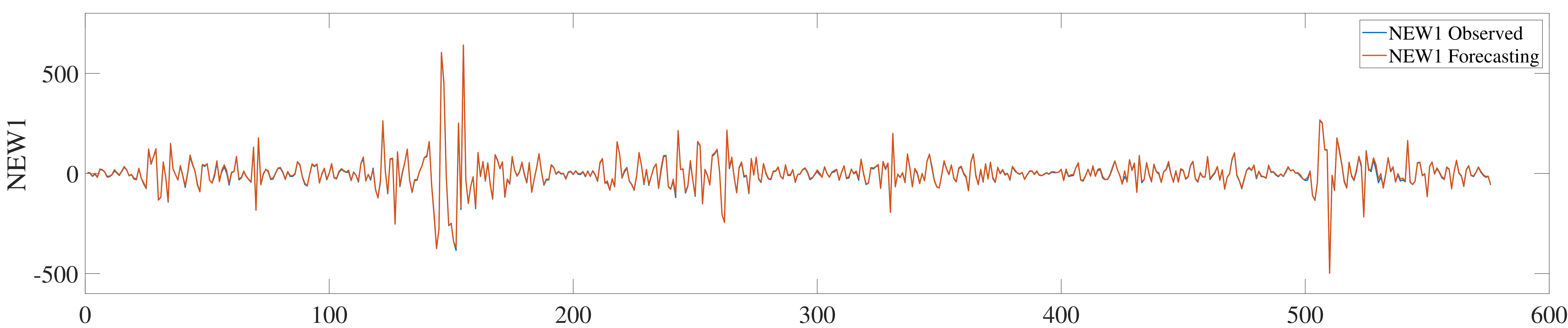}}
    \centerline{\includegraphics[width=0.5\textwidth]{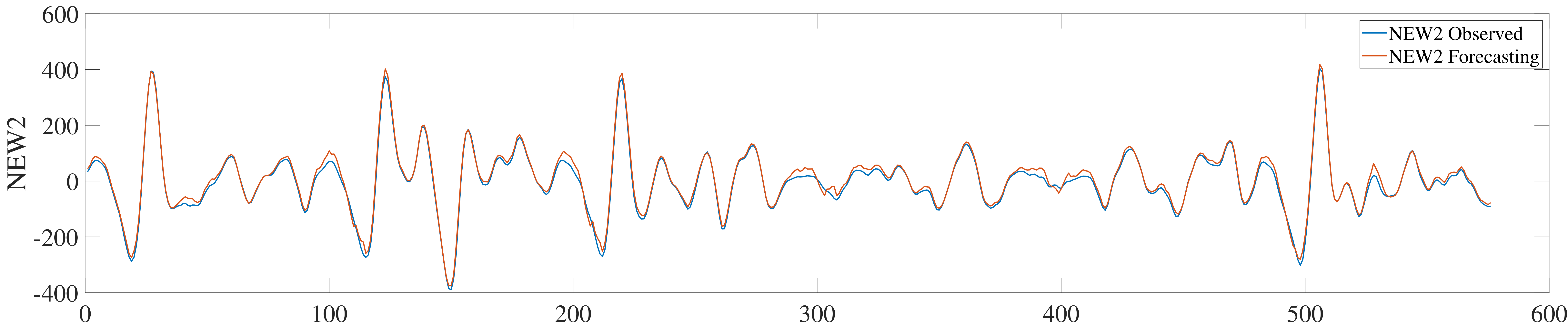}}
    \centerline{\includegraphics[width=0.5\textwidth]{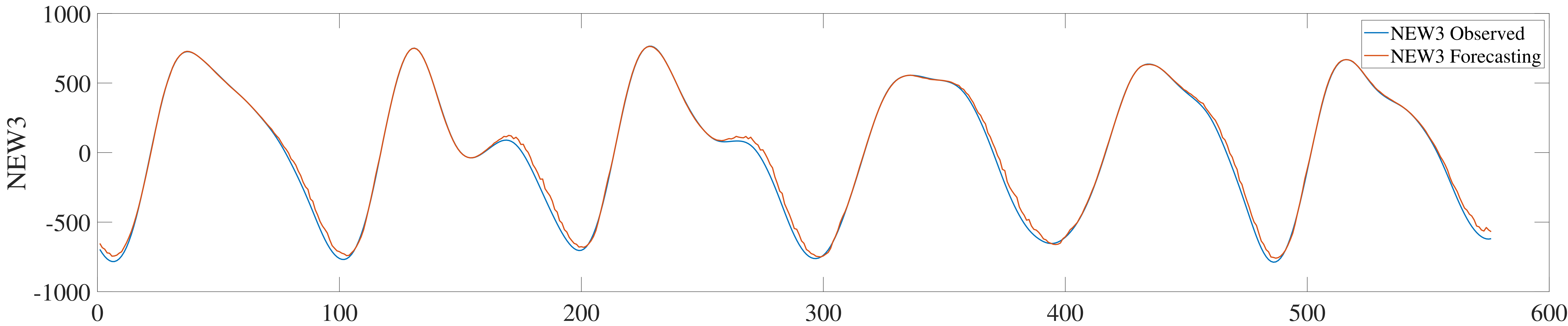}}
    \centerline{\includegraphics[width=0.5\textwidth]{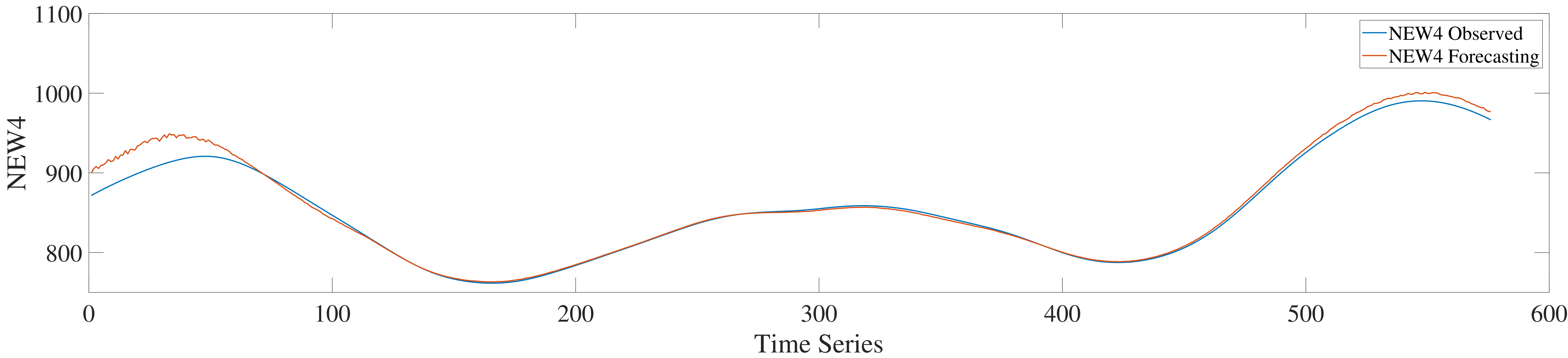}}
    \caption{The results of the Subsequences. }
    \label{fig:subsequence prediction results}
\end{figure}

%\begin{figure}[htb]
%    \centerline{\includegraphics[width=0.5\textwidth]{figs/ceemdan-se-gwo-lstm.eps}}
%    \caption{The results of the CEEMDAN-SE-GWO-LSTM. }
%    \label{fig:11}
%\end{figure}

%\begin{figure}[htb]
%    \centerline{\includegraphics[width=0.5\textwidth]{figs/oursmethod.eps}}
%    \caption{The results of This Paper. }
%    \label{fig:oursmethod}
%\end{figure}
 Fig.~\ref{fig:Comparison of different models} depicts the qualitative results of different models on part of the test set. BP network obviously suffers from outliers caused by traffic congestion. LSTM handles the traffic congestion prediction better, but it fails to capture the trend change timely. Although GWO-LSTM demonstrates good numeric performance, it has obvious fluctuations at certain times, which is caused by the non-stationary signals. Other other hand, by involving the CEEMDAN decomposition and SE reconstruction, our model's predictions can be well aligned with the observed traffic flow.

%  , it can be seen from the figure that the limitation of the BP network is that it has a significant error in the prediction of extreme values, and traffic congestion often occurs in the period when the extreme values are generated. LSTM model's predicted value can better reflect the traffic flow trend, and it performs well at extreme values. However, when the traffic flow has a long-term trend, it has a hysteresis, which cannot perfectly solve the problem of timely forecasting. GWO-LSTM shows that the error performs well in each quantization index. However, the forecast results will have large fluctuations at certain times; this is due to the characteristics of LSTM and non-stationary signals. When LSTM uses the node information of adjacent non-stationary signals, it will be affected by a considerable fluctuation value and generate a significant prediction error. 

Fig.~\ref{fig:subsequence prediction results} shows the predictions of each component reconstructed from SE, and these results are merged together to obtain the final outcome. It is obvious that the low-frequency part NEW4 has a large error due to the autocorrelation issue, which will result in an increase in the cumulative error along with the temporal dimension. This indicates the importance of spatiotemporal sparse optimization in our model to vanish such autocorrelation.

\section{Conclusion}

In this paper, we propose a robust model for short-term traffic flow prediction. By utilizing the signal decomposition, GWO optimization, LSTM model and spatiotemporal feature optimization, our method can better handle the non-stationary characteristics of short-term traffic flow, solve the over-fitting and local optimal problems , and avoid the strong autocorrelation issue. We compare our model with several state-of-the-art baselines and demonstrate the superiority of our model.

% In future work, higher-quality input features and the intrinsic information of each component should be further explored, and a better prediction model can effectively improve the prediction performance. For practical applications, it would be more practical to expand the prediction results into intervals.

%%%%%%%%%%%%%%%%%%%%%%%%%%%%%%%%%%%%%%%
\bibliographystyle{unsrt}
\bibliography{ref.bib}

\end{document}